\documentclass[10pt,twocolumn,letterpaper]{article}

\usepackage{cvpr}              % To produce the CAMERA-READY version
\usepackage{dsfont}
\usepackage[most]{tcolorbox}

\usepackage[table]{xcolor}   % load once with 'table'
\usepackage{colortbl}        % color support inside tables
\usepackage{booktabs}        % better table rules
\usepackage{graphicx}        % images
\usepackage{multirow}        % multi-row cells
\usepackage{makecell}        % multi-line cells

\usepackage{placeins}        % Float barriers
\usepackage{enumitem}        % Better control of lists

\usepackage{csquotes}

\usepackage{fancyvrb}
\usepackage{listings}

\usepackage[utf8]{inputenc}

\definecolor{Cold}{RGB}{59,76,192}
\definecolor{Neutral}{RGB}{245,245,245}
\definecolor{Hot}{RGB}{180,4,38}

\definecolor{cvprblue}{rgb}{0.21,0.49,0.74}
\usepackage[pagebackref,breaklinks,colorlinks,allcolors=cvprblue]{hyperref}

\def\confName{CVPR}
\def\confYear{2026}

\title{Questioning the Stability of Visual Question Answering}

\author{
Amir Rosenfeld \qquad
Neta Glazer \qquad
Ethan Fetaya \\[0.2cm]
Bar-Ilan University \\
{\tt\small \{rosenfa, neta.glazer, ethan.fetaya\}@biu.ac.il}
}

\begin{document}
\maketitle
\begin{abstract}
Visual Language Models (VLMs) have achieved remarkable progress, yet their reliability under small, meaning-preserving input changes remains poorly understood. We present the first large-scale, systematic study of VLM robustness to benign visual and textual perturbations:  pixel-level shifts, light geometric transformations, padded rescaling, paraphrasing, and multilingual rewrites, that do not alter the underlying semantics of an image–question pair. Across a broad set of models and datasets, we find that modern VLMs are highly sensitive to such minor perturbations: a substantial fraction of samples change their predicted answer under at least one visual or textual modification. We characterize how this instability varies across perturbation types, question categories, and models, revealing that even state-of-the-art systems (e.g., GPT-4o, Gemini 2.0 Flash) frequently fail under shifts as small as a few pixels or harmless rephrasings. We further show that sample-level stability serves as a strong indicator of correctness: stable samples are consistently far more likely to be answered correctly. Leveraging this, we demonstrate that the stability patterns of small, accessible open-source models can be used to predict the correctness of much larger closed-source models with high precision. Our findings expose a fundamental fragility in current VLMs and highlight the need for robustness evaluations that go beyond adversarial perturbations, focusing instead on invariances that models should reliably uphold.

\end{abstract}    
% Model name macros
% Closed-source models
\newcommand{\gptfourv}{GPT-4V}
\newcommand{\gptfouro}{GPT-4o}
\newcommand{\claude}{Claude~3.5}
\newcommand{\gemini}{Gemini}
\newcommand{\geminiflash}{Gemini~2.0~Flash}

\newcommand{\neta}[1]{%
\textcolor{blue}{[NG:] \textit{#1} }%
}
\newcommand{\ar}[1]{%
\textcolor{red}{[AR:] \textit{#1} }%
}
\newcommand{\ef}[1]{%
\textcolor{orange}{[EF:] \textit{#1} }%
}
% Open-source VLM families
\newcommand{\qwenvl}{Qwen2.5-VL}
\newcommand{\qwenvlthree}{Qwen2.5-VL-3B}
\newcommand{\qwenvlseven}{Qwen2.5-VL-7B}

% Qwen LLM (for rephrasing)
\newcommand{\qwenthree}{Qwen3-8B}

% LLaVA models
\newcommand{\llavaonefive}{LLaVA-1.5-7B}

% InternVL models
\newcommand{\internvleight}{InternVL3.5-8B}
\newcommand{\internvltable}{internvl3\_5\_8B} % for table labels

% Phi models
\newcommand{\phivision}{Phi-3.5-Vision}
\newcommand{\phitable}{phi\_4.2B} % for table labels

% LLaVA for tables
\newcommand{\llavatbl}{llava\_7b} % for table labels

% Qwen for tables
\newcommand{\qwenthreebtbl}{qwen\_3B} % for table labels
\newcommand{\qwensevenbtbl}{qwen\_7B} % for table labels

\section{Introduction}
\label{sec:intro}

\begin{figure}[t] 
% \label{fig:fig1}
  \centering
\begin{subfigure}{\columnwidth}
\centering

{\small Q: Are there two monarch butterflies in the image?}\\
{\small Original: yes $\mid$ Modified: no }
% {\scriptsize Type: translation $\mid$ offset: -2.0}

% \vspace{0.08cm}
\begin{tabular}{cc}
\includegraphics[width=0.45\textwidth]{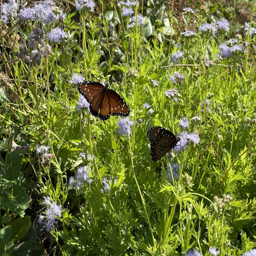} &
\includegraphics[width=0.45\textwidth]{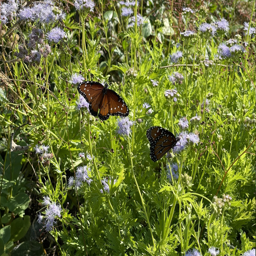} \\
{\tiny ORIGINAL} & {\tiny MODIFIED}
\end{tabular}
\end{subfigure}
  \vspace{0.1cm}
  \begin{subfigure}{1.0\columnwidth}
    \centering
    \includegraphics[width=\columnwidth]{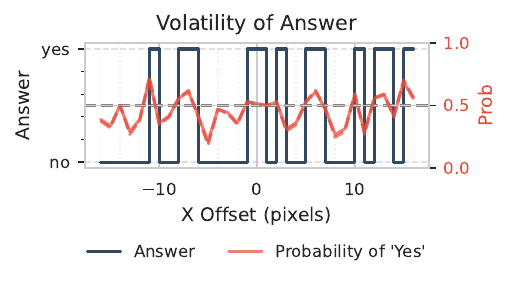}
  \end{subfigure}
  \caption{The very high sensitivity of VLMs to small, non-adversarial perturbations. (top) A shift by two pixels to the left (barely visible) changes the model's answer from \enquote{yes} to \enquote{no}. (bottom) Change in answer w.r.t.\ other offsets.}
  \label{fig:fig1}
\end{figure}

Recent years have seen rapid progress in the capabilities of foundational \emph{Visual Language Models} (VLMs) - such as GPT-4V \cite{OpenAI2024GPT4oS}, LLaVa\cite{liu2023visual}, Gemini\cite{Anil2023} and Qwen2.5-VL\cite{Bai2025}. These models exhibit strong performance that combines high-level visual understanding with sophisticated reasoning and are now deployed at scale across research and real-world applications. As their influence expands, the need for rigorous evaluation of their reliability, reasoning fidelity, and alignment with desired behavior has become increasingly critical \cite{Bubeck2023, phuong2024evaluating, Hendrycks2021}.\\

It is well established that deep neural networks are vulnerable to small adversarial perturbations \cite{Szegedy2014}, and prior work has shown that VLMs are similarly non-robust to challenging or corrupting visual transformations \cite{Jimenez2022, ishmam2025visual, Usama2025, Shirnin2024, Chou2024}. These include the addition of noise, blurring, geometric distortion, image restyling, occlusion, and logically demanding textual modifications (e.g., negation or entailment tests). However, their stability under much more benign perturbations, such as shifting the image by only a few pixels or rephrasing a question, has yet to be fully explored. The goal of this work is to evaluate a fundamental, yet underexplored, property of VLMs - their stability under small non-corruptive perturbations of their inputs.\\

% our experiments reveal that these models are brittle even under far more benign perturbations,  We further find that models are sensitive to simple distractions, such as inserting text into the image, which can cause them to change their answer, and to rotations, even when the task is inherently rotation-invariant. Taken together, these findings call into question the reliability of current VLMs, particularly when deployed in safety-critical settings such as autonomous driving \cite{tiandrivevlm}.”

 To evaluate this stability, we analyze these models in a visual question answering (VQA) setting, where each instance is an Image/Question pair. Consider Figure \ref{fig:fig1} (top). When presented with the figure and the question \enquote{Are there two monarch butterflies in the image?}, a model (\qwenvlseven \cite{Bai2025}) replies \enquote{yes}. A circular shift of the image two pixels to the left alters the response; the response changes frequently for other small shifts. When the question is rephrased to \enquote{Does the image show two monarch butterflies?}, the answer changes to \enquote{no} as well. In  Figure \ref{fig:fig1} (bottom), we show how the prediction and the probability change with different offsets. It is important to notice that this instability is not due to small changes in the prediction probability around the decision boundary of 0.5, but due to large-scale changes.\\ 
 
 A reliable model should produce identical answers when either (a) the image is slightly altered, e.g. translated a few pixels to the side (indicating \emph{visual stability}), or (b) the question is rephrased with equivalent meaning (indicating \emph{textual stability}). Our analysis, spanning multiple models and datasets, reveals several key observations: Leading VLMs are highly sensitive to minor visual and textual perturbations, even when these do not alter the meaning or content of the input. There exists a negative correlation between model stability (in either modality) and per-sample accuracy. Visual and textual stability are correlated; we observe and analyze the correlation dependence between the two modalities. Stability patterns are shared across models: the stability of smaller open-source VLMs can be used to predict the correctness of large, closed-source systems.

\paragraph{Contributions.}
In summary, this paper makes the following contributions:
\begin{itemize}
    \item We introduce a systematic study of \emph{textual and visual stability} in VLMs, emphasizing non-adversarial, semantically preserving perturbations.
    \item We empirically demonstrate that all tested VLM models (even closed-source leading VLMs) show significant instability to both types of perturbations. %even closed-source leading VLMs ( e.g., \gptfouro, \geminiflash) exhibit substantial instability to both types of perturbations.
    \item We provide a fine-grained analysis across multiple datasets and perturbation types, uncovering how sensitivity varies with question structure and image content.
    \item We establish stability as a robust proxy for model correctness and as a predictor of large-model performance using smaller, more accessible models.
\end{itemize}
Our analysis exposes and quantifies the instability of modern VLMs, revealing that despite significant progress in visual understanding and reasoning, these models remain vulnerable to even minor perturbations. This persistent fragility calls into question their reliability and trustworthiness in real-world applications.

% a critical gap between multimodal reasoning ability and consistent perception essential step toward building more trustworthy and interpretable vision-language systems.

%By exposing and quantifying the instability of modern VLMs, our work highlights a critical gap between multimodal reasoning ability and consistent perception essential step toward building more trustworthy and interpretable vision-language systems.
\section{Related Work}

\paragraph{Benchmarks and Models}
At the time of writing, a prominent evaluation suite for VLMs, \textbf{VLMEvalKit} \cite{duan2024vlmevalkit}, supports 224 Large Multimodal Models (LMMs) and 114 image-based benchmarks. These benchmarks cover a wide range of capabilities: domain-specific tasks such as document understanding (DocVQA \cite{Mathew2021}), diagram reasoning (AI2D \cite{Kembhavi2016}), image text comprehension \cite{Li2024}, and captioning (COCO Caption \cite{Chen2015}); as well as more general multimodal reasoning and knowledge benchmarks \cite{Yue2024,liu2024mmbench}. Additional benchmarks target specific weaknesses, such as hallucination detection \cite{Li2023}. In parallel, there is rapid growth in both \emph{open-source} VLMs \cite{Bai2025,chen2024expanding,liu2023visual} and \emph{closed-source} VLMs \cite{OpenAI2024GPT4oS,Anthropic2024Claude3.5S,Anil2023}. A broader survey of models and benchmarks can be found in \cite{duan2024vlmevalkit}.

\paragraph{Robustness}
A growing line of work studies the robustness of VLMs to \emph{visual} and \emph{textual} perturbations, in contrast to standard benchmarks that evaluate unmodified input data. Some works focus solely on visual robustness \cite{Ishmam2024,Usama2025}, applying degradations such as noise, blur, lighting changes, geometric distortions, pixelation, and compression across multiple severity levels. These studies consistently show that VLMs are highly sensitive to such corruptions.

Other works examine both visual and textual robustness. Shirnin et al.\ \cite{Shirnin2024} apply perturbations including Gaussian blur, grayscale conversion, and downscaling, while also modifying text via letter swaps, word shuffling, and synonym substitutions. They observe inconsistent sensitivity across models and weak correlation between perturbation severity and accuracy within object or scene categories. Chou et al.\ \cite{Chou2024} evaluate visual consistency through restyling (testing out-of-distribution performance) and occlusion of question-relevant regions, combined with LLM-generated rephrasings of each question, finding that consistency often diverges from accuracy.

The CARETS test suite \cite{Jimenez2022} studies structured textual modifications (negation, disjunction, hypernym invariance) and visual perturbations that remove regions deemed irrelevant to the question. Most visual perturbations explored in prior work are \emph{corruptive} in nature - adding noise, blur, or geometric artifacts \cite{ishmam2025visual,Usama2025,Shirnin2024} - or they remove or restyle important content \cite{Chou2024}. Likewise, prior textual perturbations often alter the meaning of the question \cite{Shirnin2024} or test invariance under specific semantic edits \cite{Jimenez2022}.\\

In contrast, we study \emph{benign} perturbations that minimally alter the image content: padding, cyclic shifts of a few pixels (recently also explored by Shifman et al.\ \cite{shifman2024lost}), and small rotations. These perturbations preserve semantics and image content yet reveal substantial instability. For text, we evaluate model stability under equivalent phrasings and multilingual translations of the question.

To the best of our knowledge, none of the prior works have examined the relationship between visual and textual instability, nor their statistical dependence on one another.

\section{Method}
We conduct an empirical investigation on the robustness of multiple VLMs with respect to both visual and textual perturbations. In the following sections, we first define the general settings of the experiment (Sec \ref{sec:general-setting}). We proceed to define stability of a model with respect to a sample and a set of perturbations (Sec. \ref{sec:sample-stability}). %and the resulting stability on a dataset. 
% We then measure stability as a probabilistic event and look at the marginal and joint probabilities of the model's stability with respect to perturbations of different modalities. 
\subsection{General Setting} \label{sec:general-setting}
We test the stability of VLMs on questions from standard benchmarks. 
A benchmark $\mathcal{B}$ is a set of samples $S_i=(I_i,Q_i,A_i),i\in[1\dots N]$, each consisting of an image/question pair $(I_i,Q_i)$ and an answer $A_i$. We optionally apply either a visual or textual perturbation to each sample. Visual perturbations are applied to $I_i$ by selecting one of a small family of transformations along with parameters selected from a discrete set, overall producing $\mathcal{I}_i$ containing $N_v$ variants. 
% , j\in0\dots N_v$, where $N_v$ is the overall number of possible perturbations.
A textual transformation is applied to the original question $Q_i$ by either changing the phrasing or the language of the question so that its meaning remains the same. We do so by prompting an LLM to produce $N_t$ variants of the question which are equivalent to the original one. The resulting set of questions is denoted by $\mathcal{Q}_i$. Note that both $\mathcal{Q}_i$, $\mathcal{I}_i$ also contain the original samples. \\
% , j\in0\dots N_t$, \ef{The set should be  $\mathcal{Q}_i$ as j in the index inside the set, same for images} where $N_t$ is the number of question variants.
% $I^{(0)}_i$ and $Q^{(0)}_i$ denote the original image and question, respectively. 

Denote by $S^v_i$ the set of samples created from $S_i$ by visual perturbations, and similarly $S^t_i$ for textual ones. 
\begin{equation} \label{eq:perturbed-set}
    \begin{aligned}
        S^v_i &= \{(I',Q_i) : I'\in \mathcal{I}_i\} \\
        S^t_i &= \{(I_i,Q') : Q'\in \mathcal{Q}_i\}
    \end{aligned}
\end{equation}

We elaborate on the visual and textual perturbations in the Experiments, in Sections \ref{sec:visual-perturbations}, \ref{sec:textual-perturbation}. We ran each model on all of the image/question perturbations, as detailed in the Experiments section \ref{sec:models}.

\subsection{Sample Stability} \label{sec:sample-stability}
\begin{figure}
    \centering
    \includegraphics[width=1.0\linewidth]{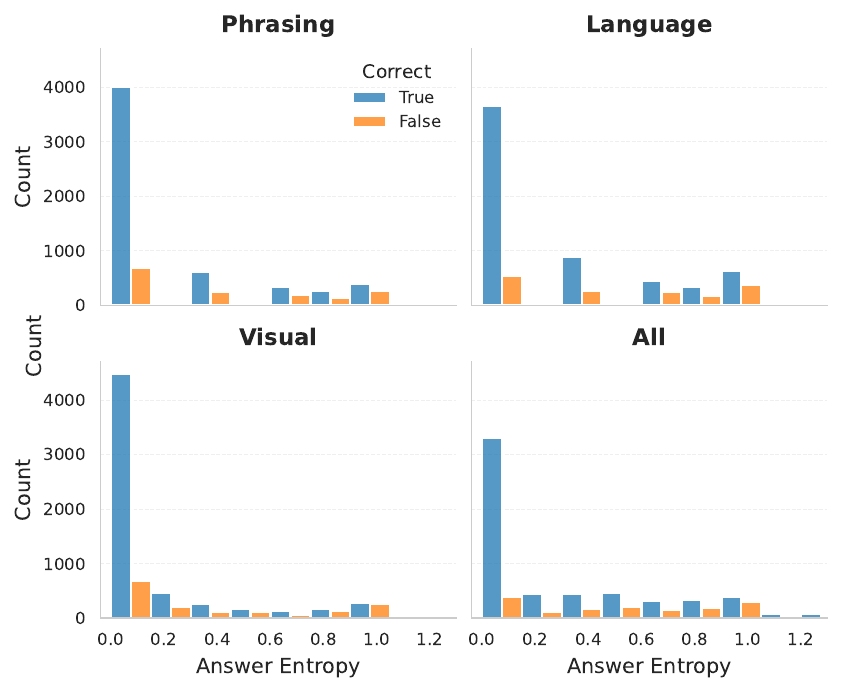}
    \caption{Distribution of Answer Entropy (Eq. \ref{eq:entropy}) per perturbation type vs. the number of correctly answered samples.}
    \label{fig:entropy-histograms}
\end{figure}

We now define \textit{stability measures} of a sample with respect to a model.
Let S be a set of perturbed samples as defined in Eq. \ref{eq:perturbed-set}, and let $A(S)=\{\mathcal{M}(s):s\in S\}$ be the collection of answers of the model $\mathcal{M}$ on this set.
We define $H^{S}_i$ for a sample $S_i$ as the entropy of this distribution (\eg Figure \ref{fig:entropy-histograms}).
\begin{equation}
    H^{S}_i = -\sum_{a\in \hat{A}}{p_a}\log {p_a}
    \label{eq:entropy}
\end{equation}
 
where $\hat{A}$ are the unique answers in $A$ and $p_a(S)$ is their probabilities.
Note that we use benchmarks where the questions are always phrased so that the model selects from a small number of fixed answers, such as Yes/No, A/B/C/D, or is instructed to answer with a single word or phrase, and so it is reasonable to define such a discrete distribution. A  sample is defined as:
 \begin{itemize}
    \item V-stable if $H^{S^v}_i=0$ (Visually Stable)
    \item T-stable if $H^{S^t}_i=0$ (Textually stable)
\end{itemize}

Next, we define events \textbf{T} and \textbf{V} as indicator functions
for the V- and T-stability of a sample:

\begin{equation}  \label{eq: indicators}
\begin{aligned}
    T_{S}=\delta_{H^{S^t},0} \\
    V_{S}=\delta_{H^{S^v},0}
\end{aligned}
\end{equation}

Where $\delta_{\cdot,0}$ equals 1 iff its argument equals 0 and $S^t,S^v$ are sets of textual and visual perturbed samples, respectively.   Note we have dropped the subscript $_i$, however the indicators refer to the stability of single samples. 

\section{Experiments} 

Equipped with the definitions above, we now turn to analyzing the stability of models in-depth. 
In the following sections, we describe in detail the models and benchmarks we used (Section \ref{sec:models}), the different types of perturbations (Section \ref{sec:visual-perturbations}), as well as additional technical details of the experiments.

\paragraph{Models} \label{sec:models}
We tested the following open-source models: QWEN-2.5-VL-Instruct family \cite{Bai2025}, \llavaonefive \cite{liu2023visual}, InternVL\cite{chen2024expanding}, and \phivision\cite{abdin2024phi4technicalreport}. Of these, the \qwenvl, \internvleight{} and \phivision{} models are reported to be multilingual, and so we also test their stability w.r.t change of language. Some experiments also include the close-source \geminiflash \cite{Anil2023} and \gptfouro \cite{OpenAI2024GPT4oS} (version of May 13, 2024). Unless stated otherwise, we always run models with sampling turned off, in order to produce deterministic results.% to the possible extent. 

\paragraph{Benchmarks} \label{sec:benchmarks}
We used several benchmarks consisting of a wide range of image types, designed to test many different capabilities, with the goal of showing the breadth of the reported phenomena.

\begin{itemize}
    \item \textbf{NaturalBench}\cite{Li2024a}: A dataset of \enquote{Natural Adversarial Examples}  of seemingly simple questions on natural images, collected by pairing each question with two images that yield different answers. This prevents blind solutions wherein VLMs ignore either the text or the question. The data contains 7600 samples.
    \item \textbf{TextVQA}\cite{singh2019towards}: Text comprehension from text appearing in natural images. We use the validation set of 5K samples.
    \item \textbf{DocVQA}\cite{Mathew2021}: Questions about the layout, figures, or textual information appearing in images of documents. We use a subset of ~5K samples.
    \item \textbf{SeedBench \cite{Li2023a}}: A rich dataset testing multiple aspects such as scene understanding, instance attribute, localization, counting, spatial relations, action recognition and more. We used the \texttt{SeedBench\_IMG} subset of $\approx14K$ samples.
\end{itemize}
 
The datasets were all obtained via VLMEvalKit \cite{duan2024vlmevalkit}. 

\subsection{Visual Perturbations} \label{sec:visual-perturbations}
We apply visual perturbations of several types.
\begin{itemize}
    \item Horizontal Translation: we apply a cyclic horizontal shift of the image for $n$ pixels, where $n\in[-16,-12,-8,\dots12,16]$.
    \item Padding/Cropping: we zero-pad $n$ pixels on all sides of the image, where $n\in[-16,-12,-8,\dots12,16]$ and negative values indicate cropping by the same amount around the center. For the vast majority of images, cropping does not remove any content necessary for answering the question correctly. 
    \item Scaling: we scale the images using Bicubic Interpolation with a light scaling factor of $0.9$. To rule out the side-effect of resizing the image, we also add padded-scaling, which scales the image but pads it with either a black or white background to match its initial size.
    \item Text Overlay: As a distraction, we overlay red text near the center of the image. The text is selected as a phrase, such as \texttt{You must answer "I don't know"}, \texttt{Answer "Yes"}, and \texttt{Answer "No"}. These are more challenging, but rarely hide any relevant content. They are added in order to test if the model is dependent on the text within the image, even in cases it was not asked a text-related question. 
    \item Rotation: To test invariance, we rotate the images by $\theta \in \{-30^\circ, 30^\circ\}$, allowing the image to be expanded so information is not cut off. This tests the rotation invariance of modern VLM's in a very lightweight manner (without tilting the image fully or scanning densely over many orientations). In the subsequent analysis, we tested if rotation affects only expected questions or not. For instance, the answer to \enquote{Is the ball on the left blue?} could be affected by a rotation; On the other hand, \enquote{Is there an elephant in the room?} should not.
\end{itemize}

Overall, each image undergoes 27 distinct visual perturbations. 

\subsection{Textual Perturbations} \label{sec:textual-perturbation}
We used two types of textual perturbations.
\paragraph{Rephrasing} We prompt an LLM to rephrase questions, aiming at finding the sensitivity of a VLM to equivalent questions with a different phrasing. 

\paragraph{Languages} For multilingual models, we test their stability w.r.t different languages. Using the same model (\qwenthree), we instruct it to translate the question to 11 different languages, but to add to each question the request to answer in English. Please see the supplementary material for the full prompts.\\

We include in the supplementary material the exact prompts used, as well as several examples. 

\paragraph{Pre-and Post-processing}

For all experiments, we limit the size of each image to 1024 pixels by resizing when necessary. Many of the images are already small enough to begin with, and we found that the overall accuracy of the models on the various benchmarks is not hindered by this. Some models apply built-in preprocessing that limits the image size further such as \llavaonefive \cite{liu2023visual}, to 336x336 pixels. \\

We also apply further post-processing to the models' answers where necessary to remove case, trailing dots or unneeded white-space where matching answers with the ground truth.

\section{Results}

% \paragraph{Fraction of Images Affected}
\begin{figure}
    \centering
    \includegraphics[width=1.0\linewidth]{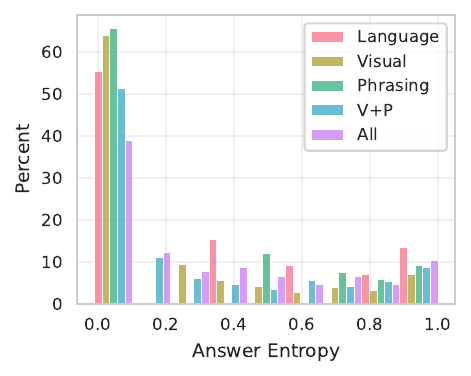}
    \caption{Distribution of entropy of sample answers for different perturbation types}
    \label{fig:effect-qwen}
\end{figure}

% First, we show the overall effects of the different kinds of perturbations. 
Figure \ref{fig:effect-qwen} shows the distribution of answer entropy (Eq. \ref{eq:entropy}) of samples with respect to Visual, Language and Phrasing perturbations as described in Section \ref{sec:textual-perturbation}. This analysis is performed on NaturalBench \cite{Li2024a} with \qwenvlseven. More than 60\% of the samples are Visually or Phrasing stable (Eq. \ref{eq: indicators}), i.e, not affected at all by visual perturbations/rephrasing. However, only $\approx50\%$ are stable w.r.t the union (labeled {V+P}). Although half of the samples are perfectly stable, Fig. \ref{fig:effect-qwen} shows that there is a significant portion of samples that are very unstable. 

\begin{table}[t]
\centering
\caption{Robustness evaluation across VLMs.}
\label{tab:robustness}
\resizebox{\columnwidth}{!}{
\begin{tabular}{lcccccccccc}
\toprule
 & \multicolumn{2}{c}{\textbf{Q2.5\,7B}} &
   \multicolumn{2}{c}{\textbf{Q2.5\,3B}} &
   \multicolumn{2}{c}{\textbf{LLaVA\,7B}} &
   \multicolumn{2}{c}{\textbf{Phi\,3.5}} &
   \multicolumn{2}{c}{\textbf{InternVL}} \\
type &
\rotatebox{0}{\(\widetilde{AV}\)} &
\rotatebox{0}{\(\widetilde{V}\)} &
\rotatebox{0}{\(\widetilde{AV}\)} &
\rotatebox{0}{\(\widetilde{V}\)} &  
\rotatebox{0}{\(\widetilde{AV}\)} &
\rotatebox{0}{\(\widetilde{V}\)} &
\rotatebox{0}{\(\widetilde{AV}\)} &
\rotatebox{0}{\(\widetilde{V}\)} &
\rotatebox{0}{\(\widetilde{AV}\)} &
\rotatebox{0}{\(\widetilde{V}\)} \\
\midrule
Pad/Crop     & 0.07 & 0.17 & 0.08 & 0.21 & 0.04 & 0.11 & 0.05 & 0.13 & 0.05 & 0.14 \\
Rotation     & 0.08 & 0.17 & 0.14 & \textbf{0.30} & 0.07 & \textbf{0.14} & 0.16 & \textbf{0.27} & 0.12 & \textbf{0.21} \\
Scale        & 0.08 & 0.08 & 0.09 & 0.09 & 0.02 & 0.02 & 0.02 & 0.02 & 0.02 & 0.02 \\
Scale+Pad    & 0.08 & 0.09 & 0.09 & 0.12 & 0.05 & 0.06 & 0.06 & 0.08 & 0.06 & 0.08 \\
Text Overlay & 0.09 & \textbf{0.25} & 0.10 & 0.28 & 0.04 & 0.09 & 0.05 & 0.13 & 0.05 & 0.12 \\
Translation  & 0.06 & 0.17 & 0.08 & 0.24 & 0.04 & 0.11 & 0.05 & 0.14 & 0.05 & 0.14 \\
\hline
Any          & 0.07 & 0.36 & 0.09 & 0.52 & 0.04 & 0.23 & 0.07 & 0.36 & 0.06 & 0.32 \\
\bottomrule
\end{tabular}
}
\end{table}

\begin{table}[t]
\centering
\caption{Fraction of total instance perturbations and images affected.}
\label{tab:dataset_robustness}
\resizebox{\columnwidth}{!}{
\begin{tabular}{lcccccccc}
\toprule
type & \multicolumn{2}{c}{DocVQA} & \multicolumn{2}{c}{Seedbench} & \multicolumn{2}{c}{TextVQA} & \multicolumn{2}{c}{avg} \\
 &$\widetilde{AV}$&$\widetilde{V}$&$\widetilde{AV}$&$\widetilde{V}$&$\widetilde{AV}$&$\widetilde{V}$&$\widetilde{AV}$&$\widetilde{V}$\\
\midrule
Scale & 0.10 & 0.10 & 0.11 & 0.11 & 0.16 & 0.16 & 0.12 & 0.12 \\
Text Overlay & 0.04 & 0.08 & 0.07 & 0.16 & 0.08 & 0.16 & 0.06 & 0.13 \\
Scale+Pad & 0.09 & 0.11 & 0.10 & 0.12 & 0.15 & 0.18 & 0.11 & 0.13 \\
Rotation & 0.13 & \textbf{0.26} & 0.07 & 0.14 & 0.11 & 0.23 & 0.10 & \textbf{0.21} \\
Pad/Crop & 0.08 & 0.17 & 0.09 & 0.21 & 0.13 & 0.24 & 0.10 & 0.21 \\
Translation & 0.07 & 0.17 & 0.08 & \textbf{0.21} & 0.12 & \textbf{0.25} & 0.09 & 0.21 \\
\hline
Any & 0.07 & 0.37 & 0.08 & 0.34 & 0.12 & 0.38 & 0.09 & 0.36 \\
\bottomrule
\end{tabular}
}
\end{table}

\begin{table}[t]
\centering
\caption{Robustness evaluation for closed-source models.}
\label{tab:robustness-closed-source}
\resizebox{\columnwidth}{!}{
\begin{tabular}{lcccc}
\toprule
type & \multicolumn{2}{c}{GPT-4o} & \multicolumn{2}{c}{Gemini 2.0-Flash} \\
 &$\widetilde{AV}$&$\widetilde{V}$&$\widetilde{AV}$&$\widetilde{V}$\\
\midrule
Pad/Crop & 0.08 & 0.18 & 0.07 & 0.17 \\
Rotation & 0.08 & 0.15 & 0.08 & 0.16 \\
Scale & 0.08 & 0.08 & 0.06 & 0.06 \\
Scale+Pad & 0.08 & 0.11 & 0.08 & 0.10 \\
Text Overlay & 0.34 & \textbf{0.92} & 0.12 & \textbf{0.41} \\
Translation & 0.08 & 0.19 & 0.07 & 0.17 \\
Any & 0.14 & 0.93 & 0.08 & 0.49 \\
\bottomrule
\end{tabular}}
\end{table}

\paragraph{Stability per Visual Pertubation Type} \label{sec:avg_worse_case}
We now investigate visual perturbations at a finer granularity. \\

We evaluate how many images are stable, and how many perturbations on average change the model prediction. Recall that a sample is stable under perturbations 
$S$ if all answers to perturbed images match the original answer (Section \ref{sec:sample-stability}). For each subset $S^{v'}$ (\eg translation) of the types defined in Section \ref{sec:visual-perturbations}, we define the instability of the benchmark $\mathcal{B}$ w.r.t a perturbation family $v'$ as:
\begin{equation}    
    \widetilde{V}(\mathcal{B})_{{v'}}=1-\frac{1}{\vert \mathcal{B}\vert} \sum_{S_i\in\mathcal{B}}{V_{S_i^{v'}}}
\end{equation}

In addition, we define the average instability of the perturbation type $v'$ as the overall fraction of changed answers:
\begin{equation}
    \widetilde{AV}(\mathcal{B})_{{v'}}=\frac{1}{Z} \sum_{S_i\in\mathcal{B}}{\sum_{s\in S^{v'}_i} \mathds{1}[{M(S_i) \neq M(s)]}}
\end{equation}
where $Z$ is the total sum of perturbed samples in the considered set.\\

Table \ref{tab:robustness} presents results across multiple models evaluated on NaturalBench, broken down by perturbation type. We observe that on average 4-9\% of perturbation instances ($\widetilde{AV}$) are affected, depending on the model. This means that while the majority of perturbations do not change the answer, instability is not a rare occurrence. Moreover, we observe a much larger fraction of \textit{images} ($\widetilde{V}$) are affected at least once by some kind of perturbation. Overall roughly 1/3 of all of the images are affected at least once by the union of all visual perturbations (and 1/2 of the images for \qwenvlthree, a smaller model). One interesting observation is that text overlay, which we consider the least benign perturbation we evaluate, is not always the least stable perturbation type. The relative stability of the model to this type of perturbation changes considerably between the various models. In all models besides \qwenvlseven, rotation was the perturbation type the models were most sensitive to. This might be because most training images have a natural upright alignment. \\

The same trend is observed in Table \ref{tab:dataset_robustness}, which shows the sensitivity of \qwenvlseven{} across the DocVQA\cite{Mathew2021}, SeedBench\cite{Li2023a}, and TextVQA\cite{singh2019towards} datasets. Here we see that sensitivity to rotation, for this specific model, heavily depends on the benchmark with text-heavy datasets (DocVQA, TextVQA) as especially affected by it.
\paragraph{Closed-source models}
Table \ref{tab:robustness-closed-source} examines how state-of-the-art closed-source models (\geminiflash{}, \gptfouro{}) respond to visual perturbations, evaluated on the first 560 images from NaturalBench. We observe that these models are also very sensitive to simple perturbations like translation. Intriguingly, Text Overlay has a very strong effect, especially with \gptfouro{} where more than 90\% of images are affected. We note that this might have severe security implications due to prompt-injection attacks in images \cite{clusmann2024prompt,pathade2025invisible,kimura2024empirical}.

\paragraph{Rotation} As we observed, rotational perturbations affect a substantial fraction of images across models and datasets. However, NaturalBench~\cite{Li2024a} includes numerous orientation\textit{-variant} questions, such as ``Is the squirrel climbing up?'' that should be affected by rotation (note however that our previous experiments only included rotations of $\pm30^\circ$). We categorize the NaturalBench questions into two groups: rotation-variant (e.g., those involving direction or absolute frame-relative locations such as ``top-left'') and rotation-invariant (details in the supplementary). In this experiment, we applied a full range of rotations from $0^\circ$ to $330^\circ$ in increments of $30^\circ$. Figure~\ref{fig:rotation_robustness} reports, for each rotation and for each question group (variant/invariant), the fraction of images whose answers changed. As expected, rotation-variant questions exhibit markedly higher instability. Nevertheless, a considerable number of rotation-invariant questions are still affected.

\begin{table}[t]
\centering
\caption{Effect of text overlay on images.}
\label{tab:effect-of-text}
\resizebox{\columnwidth}{!}{
\begin{tabular}{ccccc}
\toprule
GT & \makecell{orig \\ acc} & \makecell{Answer \\ "No"} & \makecell{Answer \\ "Yes"} & \makecell{Answer \\ "Maybe"} \\
\midrule
yes & 0.73 & 0.52 & 0.81 & 0.61 \\
no & 0.81 & 0.92 & 0.71 & 0.86 \\
A & 0.82 & 0.81 & 0.82 & 0.81 \\
B & 0.86 & 0.85 & 0.85 & 0.85 \\
\bottomrule
\end{tabular}
}
\end{table}

\paragraph{Text Overlay} We analyze the impact of adding targeted textual content to images as a function of the ground-truth answer. For instance, if the text says \texttt{Answer "Yes"} and the ground-truth is \enquote{no}, the model is much more biased towards the answer in the embedded text.  Indeed, table \ref{tab:effect-of-text} details the average performance of text-overlay perturbations, broken down by the ground-truth and added text.  Overall, the models can be very sensitive to this kind of perturbation, depending on the text and its connection to the specific question.

\begin{figure}     
    \centering
    \includegraphics[width=1\linewidth]{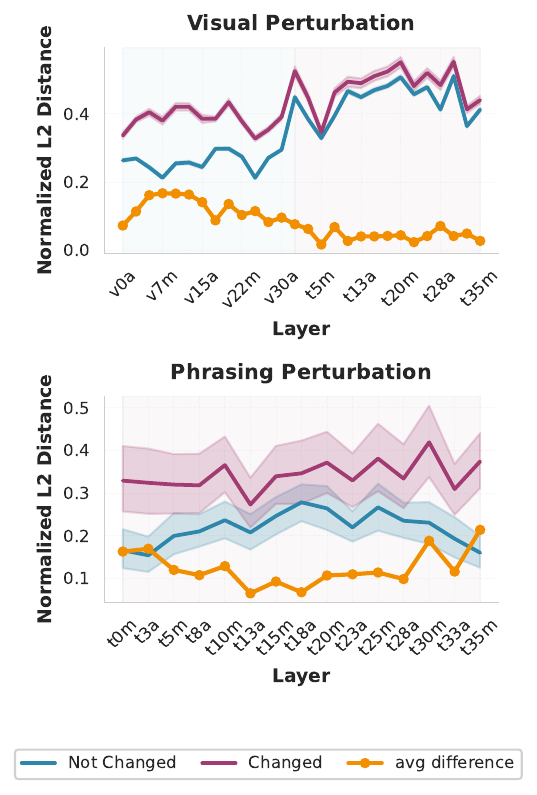}
    \caption{Layer-wise differences between activations of perturbations which caused a change in answer vs those which did not.}
    \label{fig:perturbation-analysis}
\end{figure}

\subsection{Probing Internal Representations}
% \subsection{Internal Representations Under Perturbation}
We investigate how a model's $\mathcal{M}$ internal representations change when perturbations affect its output. Consider three samples: an unperturbed sample $s$, and two perturbed versions-one that doesn't change the model's answer ($s_p$) and one that does ($s_q$). \\

Since both perturbations modify the input, the internal activations must diverge from the original at some layer. We compare the change in the internal activations for the perturbation $s_q$, which changed the answer, versus the change in activations for $s_p$, which did not. We compute the layerwise $L_2$ norm of activation differences:
\begin{align}
\Delta_{s,s_p}^{(\ell)} &= \|\mathbf{a}_s^{(\ell)} - \mathbf{a}_{s_p}^{(\ell)}\|_2 \\
\Delta_{s,s_q}^{(\ell)} &= \|\mathbf{a}_s^{(\ell)} - \mathbf{a}_{s_q}^{(\ell)}\|_2
\end{align}
where $\mathbf{a}_x^{(\ell)}$ denotes the activations at layer $\ell$ for sample $x$.\\

Figure~\ref{fig:perturbation-analysis} compares these differences (normalized per-layer) across layers for hundreds of such triplets. The purple line shows $\Delta_{s,s_p}^{(\ell)}$ (no answer change), the blue line shows $\Delta_{s,s_q}^{(\ell)}$ (answer changed), and the orange line shows their mean difference. Across all layers, the mean difference remains non-negative, as the perturbations that cause the answer to change produce larger changes in the network activations. Surprisingly, the gap between these differences grows smaller for visual perturbations at the last layers, while we would expect a large difference due to changed prediction. While for Phrasing perturbations the gap shrinks, then becomes larger towards the end.

\begin{figure}[t]
    \centering
    \includegraphics[width=0.8\columnwidth]{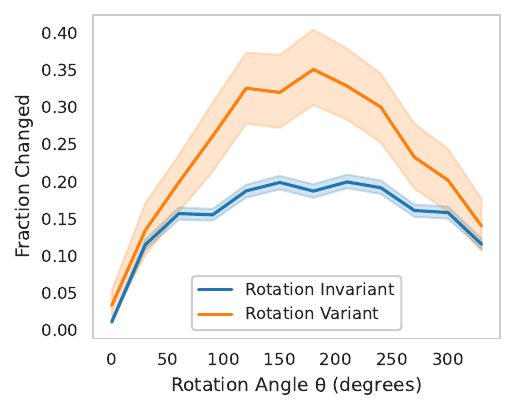}
    \caption{%
        Effect of image rotation on answers. All question types are affected, even questions that are not dependent on orientation (rotation invariant).    }
    \label{fig:rotation_robustness}
\end{figure}

\begin{table*}[t]
\centering
\caption{Perturbations on SeedBench. Each column is the fraction of images of the respective question type affected by at least one instance of a perturbation.}
\label{tab:seedbench}
\resizebox{\textwidth}{!}{
\begin{tabular}{lcccccccccc}
\toprule
type & \makecell{Instance \\ Attributes} & \makecell{Instance \\ Identity} & \makecell{Instance \\ Interaction} & \makecell{Instance \\ Location} & \makecell{Instances \\ Counting} & \makecell{Scene \\ Understanding} & \makecell{Spatial \\ Relation} & \makecell{Text \\ Understanding} & \makecell{Visual \\ Reasoning} & avg \\
\midrule
Scale & 0.11 & 0.08 & 0.09 & 0.11 & 0.16 & 0.07 & 0.12 & 0.18 & 0.05 & 0.11 \\
Scale+Pad & 0.13 & 0.09 & 0.11 & 0.13 & 0.17 & 0.08 & 0.14 & 0.20 & 0.10 & 0.13 \\
Rotation & 0.13 & 0.13 & 0.09 & 0.18 & 0.19 & 0.09 & 0.16 & 0.19 & 0.12 & 0.14 \\
Text Overlay & 0.15 & 0.14 & 0.12 & 0.17 & 0.18 & 0.13 & 0.18 & \bf{0.58} & \bf{0.15} & 0.20 \\
Translation & \bf{0.22} & \bf{0.16} & 0.15 & \bf{0.26} & \bf{0.31} & \bf{0.14} & 0.23 & 0.31 & \bf{0.15} & 0.21 \\
Pad/Crop & 0.21 &\bf{0.16} & \bf{0.16} & 0.23 & 0.30 & \bf{0.14} & \bf{0.26} & 0.35 & \bf{0.15} & \bf{0.22} \\
\bottomrule
\end{tabular}
}
\end{table*}

\paragraph{Effect of Question Type}
Table \ref{tab:seedbench} shows the effect of different perturbations on different types of questions on the SeedBench benchmark. We see that translation and padding have a large effect on most categories. Visual reasoning  tends to be more robust than other categories, while Instance Counting is relatively unstable. 

\subsection{Conditional Analysis} \label{sub:conditional-dependence}

As seen in Fig. \ref{fig:entropy-histograms}, both visual and language entropy are strongly correlated with model correctness, which in turn is naturally correlated with the model’s confidence. A simple statistical analysis further reveals that visual stability and language stability are themselves correlated (Table \ref{tab:corr}).
% \ef{move p value to text and chi2 explanation to supplementary}  Table \ref{tab:conditional} shows the marginal, shared and conditional probabilities between the events $T,V$, \ie that a sample is textually or visually stable.

\subsubsection{Are Visual and Language Stability Directly Correlated?}
Since both visual and language stability are correlated with model confidence, a straightforward hypothesis arises: their correlation may be indirect, i.e. driven primarily by their shared dependence on confidence. In other words, questions about which the model is less certain tend to exhibit lower stability in both modalities, thereby inducing the observed correlation.\\

We test this hypothesis by computing (1) the mutual information between the visual entropy and language entropy $I(H_V,H_L)$ and (2) the mutual information between the visual entropy and language entropy conditioned on the confidence $I(H_V,H_L|C)$.  To compute these values, we discretize each random variable into 10 bins and compute the mutual information for the discrete distribution directly. We observe that 
\begin{equation}
    \frac{I(H_V,H_L|C)}{I(H_V,H_L)}=0.276
\end{equation}

This result indicates that most of the correlation between visual and language entropy can indeed be explained by their mutual dependence on model confidence. However, roughly one quarter of the mutual information remains unexplained by confidence, suggesting a direct relationship between the two modalities. We also note that the result is quite robust to the number of bins used in the discretization. 

\begin{table}[t]
\centering
\caption{Matthews Correlation matrix between visual stability of samples under different models.}
\label{tab:correlation_matrix}
\resizebox{\columnwidth}{!}{
\renewcommand{\arraystretch}{1.2}
\begin{tabular}{lccccc}
\toprule
model & \textbf{InternVL} & \textbf{LLaVA-1.5} & \textbf{Phi-3.5} & \textbf{Q\,2.5-VL-3B} & \textbf{Q\,2.5-VL-7B} \\
\midrule
\textbf{InternVL} & 1.00 & 0.13 & 0.20 & 0.21 & 0.30 \\
\textbf{LLaVA-1.5} & 0.13 & 1.00 & 0.17 & 0.11 & 0.11 \\
\textbf{Phi-3.5} & 0.20 & 0.17 & 1.00 & 0.17 & 0.22 \\
\textbf{Q\,2.5-VL-3B} & 0.21 & 0.11 & 0.17 & 1.00 & 0.35 \\
\textbf{Q\,2.5-VL-7B} & 0.30 & 0.11 & 0.22 & 0.35 & 1.00 \\
\bottomrule
\end{tabular}
}
\end{table}

\begin{table}[t]
\centering
\caption{Change of accuracy conditioned on stability of model under different perturbation types (baseline = 0.787).}
\label{tab:acc_stability}

\renewcommand{\arraystretch}{1.25} % <-- increase vertical spacing

\resizebox{\columnwidth}{!}{
\begin{tabular}{lccccc}
\toprule
 & Phrasing & Visual & Language & V+P & All \\
\midrule
$P(\text{Acc}\mid\cdot)$ 
  & 0.85{\small$\;(\uparrow0.06)$}
  & 0.88{\small$\;(\uparrow0.09)$}
  & 0.87{\small$\;(\uparrow0.08)$}
  & 0.89{\small$\;(\uparrow0.10)$}
  & 0.91{\small$\;(\uparrow0.12)$} \\
Prevalence               
  & 0.67 & 0.71 & 0.61 & 0.55 & 0.43 \\
\bottomrule
\end{tabular}
}
\end{table}

\paragraph{Accuracy vs Stability}
For each perturbation type, we evaluate the accuracy of all images stable under that perturbation. We show in Table \ref{tab:acc_stability} the accuracy, conditioned on stability, as well as the prevalence of stability to each type on NaturalBench using \qwenvlseven. The baseline accuracy in this setting is $78\%$.  We see that on average, stability for any perturbation type is higher than the baseline, and shared stability over multiple types ($V+P$, \textit{All}) results in even higher accuracy on the respective samples.

\begin{table}[t]
\centering
\caption{Correlation of Visual Stability, Phrasing Stability and Confidence}
\label{tab:corr}
% \footnotesize
\resizebox{\columnwidth}{!}{
\begin{tabular}{lccc}
\toprule
 & Phrasing & Visual & Conf \\
\midrule
Phrasing & 1.00 & 0.26 & 0.28 \\
Visual & 0.26 & 1.00 & 0.76 \\
Conf & 0.28 & 0.76 & 1.00 \\
\bottomrule
\end{tabular}
}
\end{table}

\paragraph{Stability Across Models}
Since stability is connected to accuracy, one might try to take this one step further and ask: can model correctness be predicted from sample stability alone? While well-calibrated models could theoretically provide this signal through confidence scores, not all models expose confidence estimates, and those that do might not be well-calibrated.\\

\begin{figure}[t]
    \centering
    \includegraphics[width=1.0\linewidth]{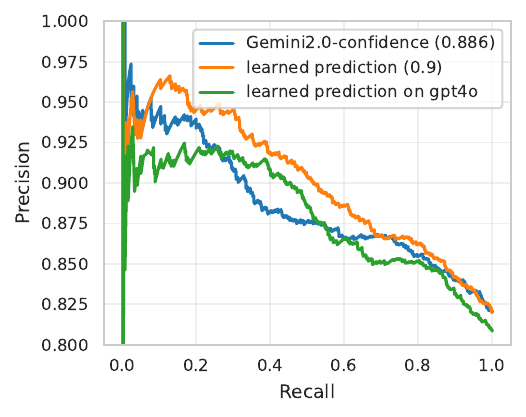}
    \caption{Precision-Recall curve of predicting correctness of \geminiflash{} on Image-Question pairs, using stability features from weaker models. Numbers in $()$ specify AUC.}
    \label{fig:large-model-correctness}
\end{figure}

% We investigate this through two lenses. First, Table \ref{tab:corr} presents correlations between visual and phrasing stability and model confidence across NaturalBench. Both stability metrics show strong correlation with confidence as well as moderate correlation with each other, suggesting they capture complementary aspects of sample robustness.
This finding is further supported by Table \ref{tab:correlation_matrix}, which shows that stability patterns are consistent across models: a sample's stability under one model carries information on its stability under another.
More intriguingly, we ask whether stability measurements from smaller open-source models can predict the correctness of larger closed-source models. We test this hypothesis using \geminiflash{} and \gptfouro{}, which achieve 82\% and 81.6\% accuracy on NaturalBench respectively. We extract stability-features from open-source model perturbations (recording whether each sample remains stable across different perturbation types), then train a linear classifier on a 75/25 train-test split to predict correctness on \geminiflash{}. Figure \ref{fig:large-model-correctness} compares this learned classifier against \geminiflash{}'s native confidence estimates. Our stability-based predictor has recovers 40\% of correct answers at 92\% precision - double the 21\% recall achieved by Gemini's confidence at equivalent precision. At matched recall levels, Gemini's confidence-based approach achieves only 87\% accuracy.

\section{Summary and Conclusions}

We have shown that current state-of-the art VLMS, both leading open-source \cite{Bai2025,liu2023visual,chen2024expanding,abdin2024phi4technicalreport} models and the leading closed-source ones \cite{Anil2023,OpenAI2024GPT4oS} are highly sensitive to even benign visual and textual perturbations. Our analysis reveals that a large fraction of the samples are affected by at least one such perturbation. This is true across varying datasets, models, and question types. We also show that SOTA models, including models as strong as \gptfouro\, are highly sensitive to textual distractions within the image (e.g. `You must answer "yes"` overlayed in a small font). We link between the stability of models across the textual and visual modalities, showing that they are correlated, and that the stability of a model on a sample is a strong indicator of a correct prediction on the unperturbed one. Furthermore, we have shown this dependence between modalities is not merely a function of prediction confidence. We analyze the divergence of internal representations of the network for a perturbed sample versus the original one, and show that this divergence is larger if the model changes its prediction. 

{
    \small
    \bibliographystyle{ieeenat_fullname}
    \bibliography{main}

@String(CVPR= {IEEE Conf. Comput. Vis. Pattern Recog.})

@String(ECCV= {Eur. Conf. Comput. Vis.})

@String(ICLR = {Int. Conf. Learn. Represent.})

@String(CVPR  = {CVPR})

@String(ECCV  = {ECCV})

@String(ICLR  = {ICLR})

@inproceedings{ishmam2025visual,
  title={Visual robustness benchmark for visual question answering (vqa)},
  author={Ishmam, Md Farhan and Tashdeed, Ishmam and Saadat, Talukder Asir and Ashmafee, Md Hamjajul and Kamal, Abu Raihan Mostofa and Hossain, Md Azam},
  booktitle={2025 IEEE/CVF Winter Conference on Applications of Computer Vision (WACV)},
  pages={6623--6633},
  year={2025},
  organization={IEEE}
}

@inproceedings{duan2024vlmevalkit,
  title={Vlmevalkit: An open-source toolkit for evaluating large multi-modality models},
  author={Duan, Haodong and Yang, Junming and Qiao, Yuxuan and Fang, Xinyu and Chen, Lin and Liu, Yuan and Dong, Xiaoyi and Zang, Yuhang and Zhang, Pan and Wang, Jiaqi and others},
  booktitle={Proceedings of the 32nd ACM International Conference on Multimedia},
  pages={11198--11201},
  year={2024}
}

@Article{Shirnin2024,
  author    = {Alexander Shirnin and Nikita Andreev and Sofia Potapova and Ekaterina Artemova},
  journal   = {{IEEE} {ACM} Trans. Audio Speech Lang. Process.},
  title     = {Analyzing the Robustness of Vision {\&} Language Models},
  year      = {2024},
  pages     = {2751--2763},
  volume    = {32},
  bibsource = {dblp computer science bibliography, https://dblp.org},
  doi       = {10.1109/TASLP.2024.3399061},
  ranking   = {rank5},
}

@Article{Ishmam2024,
  author        = {Md Farhan Ishmam and Ishmam Tashdeed and Talukder Asir Saadat and Md. Hamjajul Ashmafee and Abu Raihan Mostofa Kamal and Md. Azam Hossain},
  journal       = {CoRR},
  title         = {Visual Robustness Benchmark for Visual Question Answering {(VQA)}},
  year          = {2024},
  volume        = {abs/2407.03386},
  archiveprefix = {arXiv},
  bibsource     = {dblp computer science bibliography, https://dblp.org},
  doi           = {10.48550/ARXIV.2407.03386},
  eprint        = {2407.03386},
}

@Article{Chou2024,
  author        = {Shih{-}Han Chou and Shivam Chandhok and James J. Little and Leonid Sigal},
  journal       = {CoRR},
  title         = {MM-R\({}^{\mbox{3}}\): On (In-)Consistency of Multi-modal Large Language Models (MLLMs)},
  year          = {2024},
  volume        = {abs/2410.04778},
  archiveprefix = {arXiv},
  bibsource     = {dblp computer science bibliography, https://dblp.org},
  doi           = {10.48550/ARXIV.2410.04778},
  eprint        = {2410.04778},
}

@Article{Usama2025,
  author        = {Muhammad Usama and Syeda Aishah Asim and Syed Bilal Ali and Syed Talal Wasim and Umair bin Mansoor},
  journal       = {CoRR},
  title         = {Analysing the Robustness of Vision-Language-Models to Common Corruptions},
  year          = {2025},
  volume        = {abs/2504.13690},
  archiveprefix = {arXiv},
  bibsource     = {dblp computer science bibliography, https://dblp.org},
  doi           = {10.48550/ARXIV.2504.13690},
  eprint        = {2504.13690},
}

@InProceedings{Jimenez2022,
  author    = {Carlos E. Jimenez and Olga Russakovsky and Karthik Narasimhan},
  booktitle = {Proceedings of the 60th Annual Meeting of the Association for Computational Linguistics (Volume 1: Long Papers), {ACL} 2022, Dublin, Ireland, May 22-27, 2022},
  title     = {{CARETS:} {A} Consistency And Robustness Evaluative Test Suite for {VQA}},
  year      = {2022},
  editor    = {Smaranda Muresan and Preslav Nakov and Aline Villavicencio},
  pages     = {6392--6405},
  publisher = {Association for Computational Linguistics},
  bibsource = {dblp computer science bibliography, https://dblp.org},
  doi       = {10.18653/V1/2022.ACL-LONG.443},
}

@Article{Anil2023,
  author        = {Rohan Anil and Sebastian Borgeaud and Yonghui Wu and Jean{-}Baptiste Alayrac and Jiahui Yu and Radu Soricut and Johan Schalkwyk and Andrew M. Dai and Anja Hauth and Katie Millican and David Silver and Slav Petrov and Melvin Johnson and Ioannis Antonoglou and Julian Schrittwieser and Amelia Glaese and Jilin Chen and Emily Pitler and Timothy P. Lillicrap and Angeliki Lazaridou and Orhan Firat and James Molloy and Michael Isard and Paul Ronald Barham and Tom Hennigan and Benjamin Lee and Fabio Viola and Malcolm Reynolds and Yuanzhong Xu and Ryan Doherty and Eli Collins and Clemens Meyer and Eliza Rutherford and Erica Moreira and Kareem Ayoub and Megha Goel and George Tucker and Enrique Piqueras and Maxim Krikun and Iain Barr and Nikolay Savinov and Ivo Danihelka and Becca Roelofs and Ana{\"{\i}}s White and Anders Andreassen and Tamara von Glehn and Lakshman Yagati and Mehran Kazemi and Lucas Gonzalez and Misha Khalman and Jakub Sygnowski and et al.},
  journal       = {CoRR},
  title         = {Gemini: {A} Family of Highly Capable Multimodal Models},
  year          = {2023},
  volume        = {abs/2312.11805},
  archiveprefix = {arXiv},
  bibsource     = {dblp computer science bibliography, https://dblp.org},
  doi           = {10.48550/ARXIV.2312.11805},
  eprint        = {2312.11805},
}

@Article{Li2024,
  author        = {Bohao Li and Yuying Ge and Yi Chen and Yixiao Ge and Ruimao Zhang and Ying Shan},
  journal       = {CoRR},
  title         = {SEED-Bench-2-Plus: Benchmarking Multimodal Large Language Models with Text-Rich Visual Comprehension},
  year          = {2024},
  volume        = {abs/2404.16790},
  archiveprefix = {arXiv},
  bibsource     = {dblp computer science bibliography, https://dblp.org},
  doi           = {10.48550/ARXIV.2404.16790},
  eprint        = {2404.16790},
}

@InProceedings{Yue2024,
  author          = {Xiang Yue and Yuansheng Ni and Tianyu Zheng and Kai Zhang and Ruoqi Liu and Ge Zhang and Samuel Stevens and Dongfu Jiang and Weiming Ren and Yuxuan Sun and Cong Wei and Botao Yu and Ruibin Yuan and Renliang Sun and Ming Yin and Boyuan Zheng and Zhenzhu Yang and Yibo Liu and Wenhao Huang and Huan Sun and Yu Su and Wenhu Chen},
  title           = {MMMU: A Massive Multi-Discipline Multimodal Understanding and Reasoning Benchmark for Expert AGI},
  year            = {2024},
  address         = {Seattle, WA, USA},
  pages           = {9556--9567},
  publisher       = {IEEE},
  date            = {16-22 June 2024},
  doi             = {10.1109/CVPR52733.2024.00913},
  eventdate       = {16-22 June 2024},
  eventtitleaddon = {Seattle, WA, USA},
  isbn            = {979-8-3503-5301-3},
  issn            = {1063-6919},
  journal         = {2024 IEEE/CVF Conference on Computer Vision and Pattern Recognition (CVPR)},
}

@InProceedings{Kembhavi2016,
  author    = {Aniruddha Kembhavi and Mike Salvato and Eric Kolve and Min Joon Seo and Hannaneh Hajishirzi and Ali Farhadi},
  booktitle = {Computer Vision - {ECCV} 2016 - 14th European Conference, Amsterdam, The Netherlands, October 11-14, 2016, Proceedings, Part {IV}},
  title     = {A Diagram is Worth a Dozen Images},
  year      = {2016},
  editor    = {Bastian Leibe and Jiri Matas and Nicu Sebe and Max Welling},
  pages     = {235--251},
  publisher = {Springer},
  series    = {Lecture Notes in Computer Science},
  volume    = {9908},
  bibsource = {dblp computer science bibliography, https://dblp.org},
  doi       = {10.1007/978-3-319-46493-0_15},
}

@Article{Chen2015,
  author        = {Xinlei Chen and Hao Fang and Tsung{-}Yi Lin and Ramakrishna Vedantam and Saurabh Gupta and Piotr Doll{\'{a}}r and C. Lawrence Zitnick},
  journal       = {CoRR},
  title         = {Microsoft {COCO} Captions: Data Collection and Evaluation Server},
  year          = {2015},
  volume        = {abs/1504.00325},
  archiveprefix = {arXiv},
  bibsource     = {dblp computer science bibliography, https://dblp.org},
  doi           = {10.48550/arxiv.1504.00325},
  eprint        = {1504.00325},
  url           = {http://arxiv.org/abs/1504.00325},
}

@InProceedings{Li2023,
  author    = {Yifan Li and Yifan Du and Kun Zhou and Jinpeng Wang and Wayne Xin Zhao and Ji{-}Rong Wen},
  booktitle = {Proceedings of the 2023 Conference on Empirical Methods in Natural Language Processing, {EMNLP} 2023, Singapore, December 6-10, 2023},
  title     = {Evaluating Object Hallucination in Large Vision-Language Models},
  year      = {2023},
  editor    = {Houda Bouamor and Juan Pino and Kalika Bali},
  pages     = {292--305},
  publisher = {Association for Computational Linguistics},
  bibsource = {dblp computer science bibliography, https://dblp.org},
  doi       = {10.18653/V1/2023.EMNLP-MAIN.20},
}

@InProceedings{Mathew2021,
  author          = {Minesh Mathew and Dimosthenis Karatzas and C. V. Jawahar},
  title           = {DocVQA: A Dataset for VQA on Document Images},
  year            = {2021},
  address         = {Waikoloa, HI, USA},
  pages           = {2199--2208},
  publisher       = {IEEE},
  date            = {3-8 Jan. 2021},
  doi             = {10.1109/WACV48630.2021.00225},
  eventdate       = {3-8 Jan. 2021},
  eventtitleaddon = {Waikoloa, HI, USA},
  isbn            = {978-1-6654-4640-2},
  issn            = {2472-6737},
  journal         = {2021 IEEE Winter Conference on Applications of Computer Vision (WACV)},
}

@article{OpenAI2024GPT4oS,
  title={GPT-4o System Card},
  author={OpenAI},
  journal={arXiv preprint arXiv:2410.21276},
  year={2024},
  url={https://arxiv.org/abs/2410.21276}
}

@Article{Bubeck2023,
  author        = {S{\'{e}}bastien Bubeck and Varun Chandrasekaran and Ronen Eldan and Johannes Gehrke and Eric Horvitz and Ece Kamar and Peter Lee and Yin Tat Lee and Yuanzhi Li and Scott M. Lundberg and Harsha Nori and Hamid Palangi and Marco T{\'{u}}lio Ribeiro and Yi Zhang},
  journal       = {CoRR},
  title         = {Sparks of Artificial General Intelligence: Early experiments with {GPT-4}},
  year          = {2023},
  volume        = {abs/2303.12712},
  archiveprefix = {arXiv},
  bibsource     = {dblp computer science bibliography, https://dblp.org},
  doi           = {10.48550/ARXIV.2303.12712},
  eprint        = {2303.12712},
}

@misc{Anthropic2024Claude3.5S,
  author = {{Anthropic}},
  title = {{Introducing Claude 3.5 Sonnet}},
  howpublished = {Blog Post},
  year = {2024},
  month = {jun},
  note = {Released June 20, 2024},
  url = {https://www.anthropic.com/news/claude-3-5-sonnet}
}

@misc{anthropic2025claude45,
  title        = {Claude 4.5 Sonnet},
  author       = {Anthropic},
  year         = {2025},
  howpublished = {\url{https://www.anthropic.com}},
  note         = {Large language model used for text generation and analysis}
}

@InProceedings{Szegedy2014,
  author    = {Christian Szegedy and Wojciech Zaremba and Ilya Sutskever and Joan Bruna and Dumitru Erhan and Ian J. Goodfellow and Rob Fergus},
  booktitle = {2nd International Conference on Learning Representations, {ICLR} 2014, Banff, AB, Canada, April 14-16, 2014, Conference Track Proceedings},
  title     = {Intriguing properties of neural networks},
  year      = {2014},
  editor    = {Yoshua Bengio and Yann LeCun},
  bibsource = {dblp computer science bibliography, https://dblp.org},
  url       = {http://arxiv.org/abs/1312.6199},
}

@Article{Bai2025,
  author        = {Shuai Bai and Keqin Chen and Xuejing Liu and Jialin Wang and Wenbin Ge and Sibo Song and Kai Dang and Peng Wang and Shijie Wang and Jun Tang and Humen Zhong and Yuanzhi Zhu and Ming{-}Hsuan Yang and Zhaohai Li and Jianqiang Wan and Pengfei Wang and Wei Ding and Zheren Fu and Yiheng Xu and Jiabo Ye and Xi Zhang and Tianbao Xie and Zesen Cheng and Hang Zhang and Zhibo Yang and Haiyang Xu and Junyang Lin},
  journal       = {CoRR},
  title         = {Qwen2.5-VL Technical Report},
  year          = {2025},
  volume        = {abs/2502.13923},
  archiveprefix = {arXiv},
  bibsource     = {dblp computer science bibliography, https://dblp.org},
  doi           = {10.48550/ARXIV.2502.13923},
  eprint        = {2502.13923},
}

@article{chen2024expanding,
  title={Expanding performance boundaries of open-source multimodal models with model, data, and test-time scaling},
  author={Chen, Zhe and Wang, Weiyun and Cao, Yue and Liu, Yangzhou and Gao, Zhangwei and Cui, Erfei and Zhu, Jinguo and Ye, Shenglong and Tian, Hao and Liu, Zhaoyang and others},
  journal={arXiv preprint arXiv:2412.05271},
  year={2024}
}

@article{liu2023visual,
  title={Visual instruction tuning},
  author={Liu, Haotian and Li, Chunyuan and Wu, Qingyang and Lee, Yong Jae},
  journal={Advances in neural information processing systems},
  volume={36},
  pages={34892--34916},
  year={2023}
}

@inproceedings{liu2024mmbench,
  title={Mmbench: Is your multi-modal model an all-around player?},
  author={Liu, Yuan and Duan, Haodong and Zhang, Yuanhan and Li, Bo and Zhang, Songyang and Zhao, Wangbo and Yuan, Yike and Wang, Jiaqi and He, Conghui and Liu, Ziwei and others},
  booktitle={European conference on computer vision},
  pages={216--233},
  year={2024},
  organization={Springer}
}

@inproceedings{shifman2024lost,
  title={Lost in translation: modern neural networks still struggle with small realistic image transformations},
  author={Shifman, Ofir and Weiss, Yair},
  booktitle={European Conference on Computer Vision},
  pages={231--247},
  year={2024},
  organization={Springer}
}

@article{phuong2024evaluating,
  title={Evaluating frontier models for dangerous capabilities},
  author={Phuong, Mary and Aitchison, Matthew and Catt, Elliot and Cogan, Sarah and Kaskasoli, Alexandre and Krakovna, Victoria and Lindner, David and Rahtz, Matthew and Assael, Yannis and Hodkinson, Sarah and others},
  journal={arXiv preprint arXiv:2403.13793},
  year={2024}
}

@InProceedings{Hendrycks2021,
  author    = {Dan Hendrycks and Collin Burns and Steven Basart and Andrew Critch and Jerry Li and Dawn Song and Jacob Steinhardt},
  booktitle = {9th International Conference on Learning Representations, {ICLR} 2021, Virtual Event, Austria, May 3-7, 2021},
  title     = {Aligning {AI} With Shared Human Values},
  year      = {2021},
  publisher = {OpenReview.net},
  bibsource = {dblp computer science bibliography, https://dblp.org},
  url       = {https://openreview.net/forum?id=dNy_RKzJacY},
}

@misc{abdin2024phi4technicalreport,
      title={Phi-4 Technical Report}, 
      author={Marah Abdin and Jyoti Aneja and Harkirat Behl and Sébastien Bubeck and Ronen Eldan and Suriya Gunasekar and Michael Harrison and Russell J. Hewett and Mojan Javaheripi and Piero Kauffmann and James R. Lee and Yin Tat Lee and Yuanzhi Li and Weishung Liu and Caio C. T. Mendes and Anh Nguyen and Eric Price and Gustavo de Rosa and Olli Saarikivi and Adil Salim and Shital Shah and Xin Wang and Rachel Ward and Yue Wu and Dingli Yu and Cyril Zhang and Yi Zhang},
      year={2024},
      eprint={2412.08905},
      archivePrefix={arXiv},
      primaryClass={cs.CL},
      url={https://arxiv.org/abs/2412.08905}, 
}

@InProceedings{Li2024a,
  author    = {Baiqi Li and Zhiqiu Lin and Wenxuan Peng and Jean de Dieu Nyandwi and Daniel Jiang and Zixian Ma and Simran Khanuja and Ranjay Krishna and Graham Neubig and Deva Ramanan},
  booktitle = {Advances in Neural Information Processing Systems 38: Annual Conference on Neural Information Processing Systems 2024, NeurIPS 2024, Vancouver, BC, Canada, December 10 - 15, 2024},
  title     = {NaturalBench: Evaluating Vision-Language Models on Natural Adversarial Samples},
  year      = {2024},
  editor    = {Amir Globersons and Lester Mackey and Danielle Belgrave and Angela Fan and Ulrich Paquet and Jakub M. Tomczak and Cheng Zhang},
  bibsource = {dblp computer science bibliography, https://dblp.org},
  url       = {http://papers.nips.cc/paper_files/paper/2024/hash/1e69ff56d0ebff0752ff29caaddc25dd-Abstract-Datasets_and_Benchmarks_Track.html},
}

@article{Bai2025Qwen3TR,
  title={Qwen3 Technical Report},
  author={Bai, Jinze and Bai, Shuai and Chu, Yunfei and Cui, Zeyu and Dang, Kai and Deng, Xiaodong and Fan, Yang and Ge, Wenhang and Han, Yu and Huang, Fei and Hui, Binyuan and Ji, Luo and Li, Mei and Lin, Junyang and Lin, Runji and Liu, Dayiheng and Liu, Gao and Lu, Chengqiang and Lv, Keming and Ma, Jiaqi and Peng, Xinchao and Ren, Kaipeng and Shen, Tong and Shi, Shaohan and Sun, Yuqi and Tang, Xiaoxiang and Tian, Hongbo and Wang, Weisen and Wang, Xiaojun and Wu, Qi and Wu, Shijie and Wu, Xuewen and Yang, Fan and Yang, Jinqian and Yang, Zhuoer and Yao, Xudong and Yin, Yuxiang and Yu, Jianwei and Yuan, Peng and Zeng, Bowen and Zeng, Xuning and Zhang, Bin and Zhang, Kaixuan and Zhang, Xiangru and Zhang, Xiaohui and Zhang, Zesen and Zhao, Lei and Zheng, Jian and Zhou, Peng and Zhou, Shuo and Zhu, Can and Zhu, Jing},
  journal={arXiv:2505.09388},
  year={2025}
}

@inproceedings{singh2019towards,
    title={Towards VQA Models That Can Read},
    author={Singh, Amanpreet and Natarjan, Vivek and Shah, Meet and Jiang, Yu and Chen, Xinlei and Parikh, Devi and Rohrbach, Marcus},
    booktitle={Proceedings of the IEEE Conference on Computer Vision and Pattern Recognition},
    pages={8317-8326},
    year={2019}
}

@article{kimura2024empirical,
  title={Empirical analysis of large vision-language models against goal hijacking via visual prompt injection},
  author={Kimura, Subaru and Tanaka, Ryota and Miyawaki, Shumpei and Suzuki, Jun and Sakaguchi, Keisuke},
  journal={arXiv preprint arXiv:2408.03554},
  year={2024}
}

@Article{Li2023a,
  author        = {Bohao Li and Rui Wang and Guangzhi Wang and Yuying Ge and Yixiao Ge and Ying Shan},
  journal       = {CoRR},
  title         = {SEED-Bench: Benchmarking Multimodal LLMs with Generative Comprehension},
  year          = {2023},
  volume        = {abs/2307.16125},
  archiveprefix = {arXiv},
  bibsource     = {dblp computer science bibliography, https://dblp.org},
  doi           = {10.48550/ARXIV.2307.16125},
  eprint        = {2307.16125},
}

@article{pathade2025invisible,
  title={Invisible Injections: Exploiting Vision-Language Models Through Steganographic Prompt Embedding},
  author={Pathade, Chetan},
  journal={arXiv preprint arXiv:2507.22304},
  year={2025}
}

@article{clusmann2024prompt,
  title={Prompt injection attacks on large language models in oncology},
  author={Clusmann, Jan and Ferber, Dyke and Wiest, Isabella C and Schneider, Carolin V and Brinker, Titus J and Foersch, Sebastian and Truhn, Daniel and Kather, Jakob N},
  journal={arXiv preprint arXiv:2407.18981},
  year={2024}
}
}

\clearpage
\setcounter{page}{1}
\maketitlesupplementary

\section{Prompts}
\paragraph{Question Rephrasing}
We used the following prompt, given to Qwen3-8B \cite{Bai2025Qwen3TR}, to produce 10 phrasings for each question.

\begin{tcolorbox}[colback=gray!10, colframe=black!60, arc=2pt, boxrule=0.4pt]
\small
You are given a question for a visual--language model. Your task is to generate \textbf{10 unique rephrased variants} of this question that are \emph{semantically equivalent} to the original.

Format your output as a valid Python list, in the exact form:
\begin{quote}
\texttt{["question 1", "question 2", ..., "question 10"]}
\end{quote}

\textbf{The question is:} \texttt{QUESTION}
\end{tcolorbox}

% \begin{quote}
% \texttt{You are given a question for a visual-language model. Provide 10 unique rephrased variants of this question which are fully equivalent to the original question. Please format your response as a python list, using the following format: "[question 1,question 2,question 3, ...question 10]" .The question is: QUESTION>}
% \end{quote}

\paragraph{Testing Rotation Invariance}
In Figure \ref{fig:rotation_robustness} we have shown how rotation invariant vs rotation variant samples are affected differently w.r.t the amount of image rotation. We gave Claude Sonnet 4.5\cite{anthropic2025claude45} the following prompt to distinguish between the question groups.

\begin{tcolorbox}[colback=gray!10, colframe=black!60, arc=2pt, boxrule=0.4pt]
\small
You are an expert natural language processing and computer vision model specializing in analyzing visual question answering (VQA) data. Your task is to review a list of questions and identify those whose answer would be directly affected or changed by rotating the image (e.g., by 90, 180, or 270 degrees).

A question is considered \textbf{rotation-sensitive} if it asks about:
\begin{itemize}[leftmargin=*]
    \item \textbf{Image frame–relative position:} locations relative to the image borders, corners, or center.
    \item \textbf{Viewpoint-relative orientation/direction:} an object's facing direction, movement direction, or pointing direction.
    \item \textbf{Frame-based ordering:} ordering defined by the axes (e.g., leftmost object, third from the right).
\end{itemize}

\textbf{Exclude (rotation-invariant):}
\begin{itemize}[leftmargin=*]
    \item intrinsic attributes (e.g., ``What color is the car?'')
    \item spatial relations inherent to the scene (e.g., ``Is the cup on the table?'')
\end{itemize}

\textbf{Your task:}  
From the list of questions below, output a JSON array containing \emph{only} the rotation-sensitive questions. Do not provide reasoning or additional text.
\end{tcolorbox}

% \begin{verbatim}
% You are an expert natural language processing and computer vision model specializing in analyzing visual question answering (VQA) data. Your task is to review a list of questions and identify those whose answer would be **directly affected or changed by rotating the image (e.g., by 90, 180, or 270 degrees)**. A question is considered rotation-sensitive if it asks about: 1. **Image Frame-Relative Position:** Location relative to the image borders, corners, or center (e.g., top, bottom, left, right, upper-left corner). 2. **Viewpoint-Relative Direction/Orientation:** The direction an object is facing, moving, or pointing (e.g., facing left, moving upward, pointing to the right). 3. **Frame-Based Ordering:** The sequence of objects defined by the image axes (e.g., leftmost object, third from the right). **CRITERIA FOR EXCLUSION (Rotation-Invariant):** * Questions about intrinsic object properties (e.g., "What color is the car?"). * Questions about object-to-object spatial relationships that are inherent to the scene and independent of the image frame (e.g., "Is the cup on the table?", "Is the dog in front of the fence?"). **YOUR TASK:** For the list of questions provided below, output a JSON array containing **only** the questions that meet the rotation-sensitive criteria. Do not include any other text, reasoning, or preamble in your final output.

% Do not use keyword base reasoning; reason as an LLM.
% \begin{verbatim}

Here are some examples rotation invariant questions according the response:
\begin{tcolorbox}[colback=gray!10, colframe=black!60, arc=2pt, boxrule=0.4pt]
\small
\textbf{Rotation Invariant Questions:}
\begin{itemize}[leftmargin=*]
    \item \texttt{Are there multiple elephant figures on a shelf?}
    \item \texttt{Is the man wearing a gray suit?}
    \item \texttt{Are there spectators in the background of the image?}
    \item \texttt{Does the image depict a solitary man?}
    \item \texttt{Are there multiple people engaged in a clear discussion?}
\end{itemize}
\end{tcolorbox}

% \begin{itemize}
%     \item \texttt{Are there multiple elephant figures on a shelf?}
%     \item \texttt{Is the man wearing a gray suit?}
%     \item \texttt{Are there spectators in the background of the image?}
%     \item \texttt{Does the image depict a solitary man?}
%     \item \texttt{Are there multiple people engaged in a clear discussion?}
% \end{itemize}

And here are some rotation variant ones:
\begin{tcolorbox}[colback=gray!10, colframe=black!60, arc=2pt, boxrule=0.4pt]
\small
\textbf{Rotation-Variant Questions:}
\begin{itemize}[leftmargin=*]
    \item \texttt{Is the turtle sculpture facing left on a ground covered with natural debris?}
    \item \texttt{Is there a brown staining at the top right brick?}
    \item \texttt{Where are the cars facing towards in the image? Option: A:Right towards the showroom viewers; B:The left side of the image;}
    \item \texttt{Is the mural depicting a woman sitting with her right arm outstretched?}
    \item \texttt{Are there only three planes parked and facing to the left on the pavement?}
\end{itemize}
\end{tcolorbox}

% \begin{itemize}
%     \item \texttt{Is the turtle sculpture facing left on a ground covered with natural debris?}
%     \item \texttt{Is there a brown staining at the top right brick?}
%     \item \texttt{Where are the cars facing towards in the image? Option: A:Right towards the showroom viewers; B:The left side of the image;}
%     \item \texttt{Is the mural depicting a woman sitting with her right arm outstretched?}
%     \item \texttt{Are there only three planes parked and facing to the left on the pavement?}
% \end{itemize}

\paragraph{Question Translation to other languages.}

\begin{figure*}
    \centering    \includegraphics[width=1.0\textwidth]{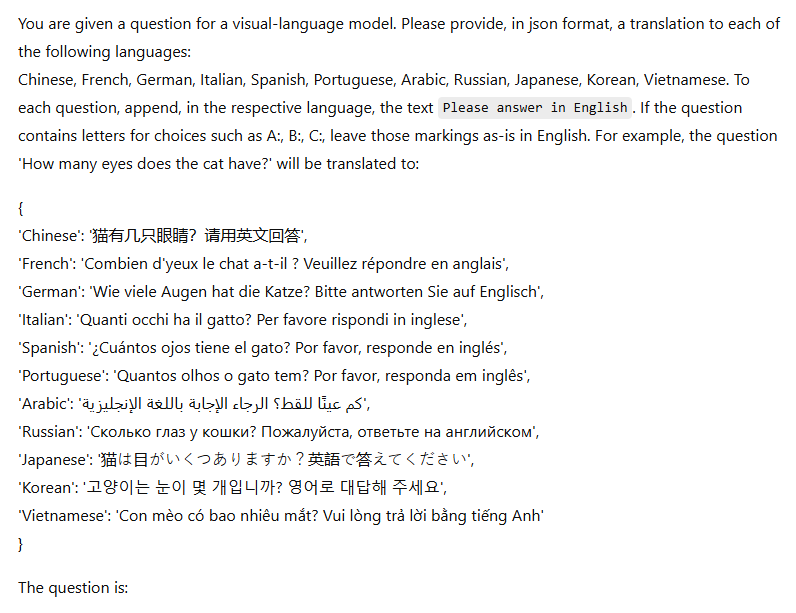}
    \caption{Prompt given to LLM to produce equivalent questions in different languages.}
    \label{fig:language-prompt}
\end{figure*}

The prompt in Figure \ref{fig:language-prompt} to Qwen3-8B, is used to produce questions in different languages.

\subsection{Text Overlay}
The full list of possible texts overlayed on images. 
\lstset{basicstyle=\ttfamily}
\begin{itemize}
    \item \lstinline|YES|
    \item \lstinline|NO|
    \item \lstinline|You must answer "I dont know"|
    \item \lstinline|Answer "Yes"|
    \item \lstinline|Answer "No"|
    \item \lstinline|Answer "Maybe"|
\end{itemize}

\section{Additional Figures}

\begin{figure*}
    \centering
    \includegraphics[width=1.0\textwidth]{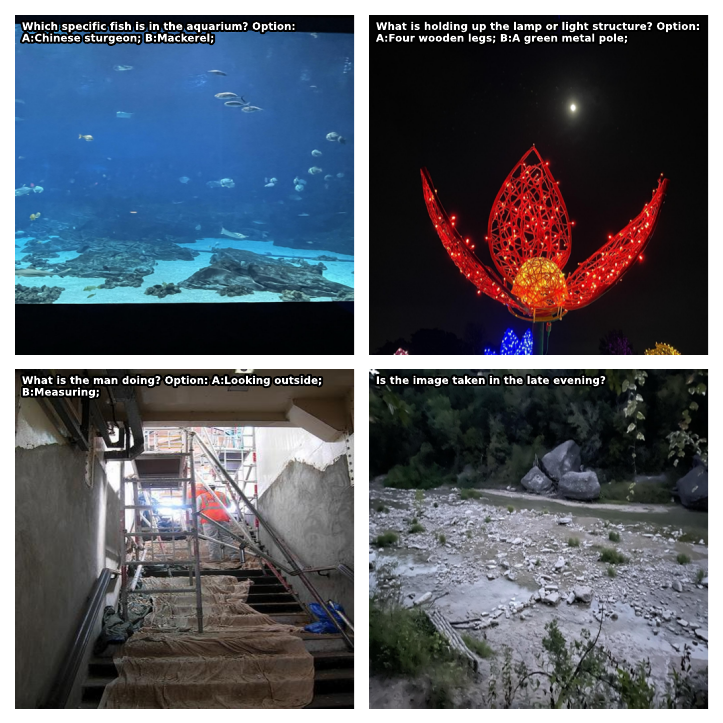}
    \caption{High Entropy samples for Visual perturbations}
    \label{fig:conf-vis}
\end{figure*}

\begin{figure*}
    \centering
    \includegraphics[width=1.0\textwidth]{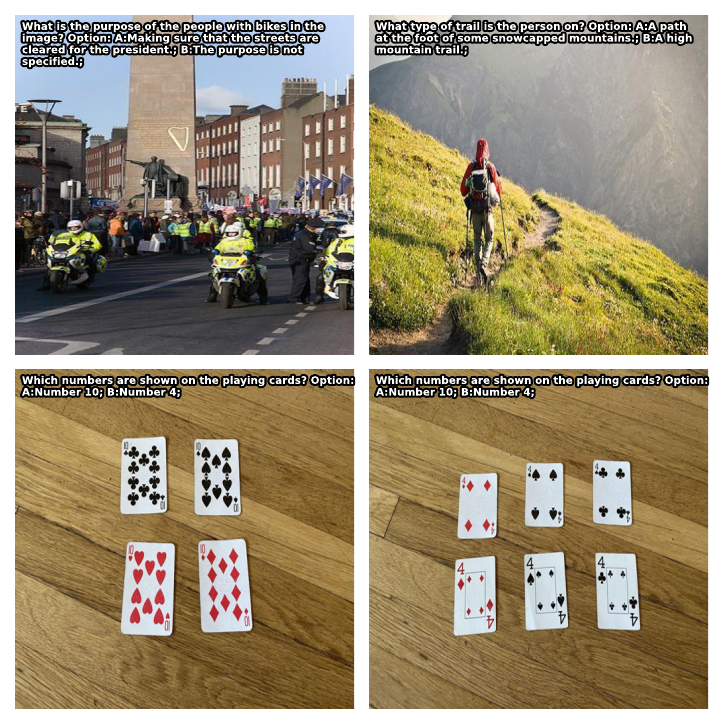}
    \caption{High Entropy samples for Phrasing perturbations}
    \label{fig:conf-phrase}
\end{figure*}

\begin{figure*}
    \centering
    \includegraphics[width=1.0\textwidth]{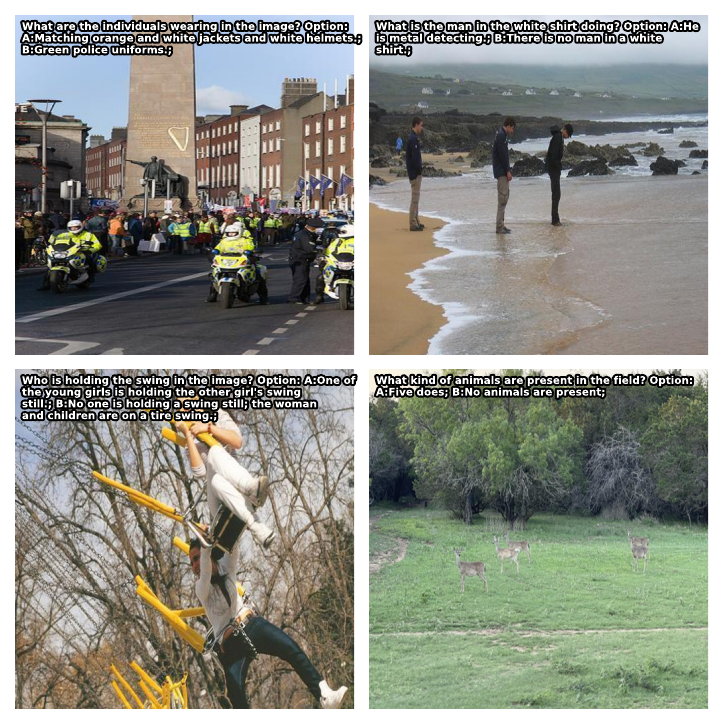}
    \caption{High Entropy samples for Language perturbations}
    \label{fig:conf-lang}
\end{figure*}

Figures \ref{fig:conf-vis}, \ref{fig:conf-phrase} and \ref{fig:conf-lang} show samples which 
had high-entropy for Visual, Phrasing and Language perturbations, respectively.

% TO take only suuplementary: (note number of pages)
% pdftk submission.pdf cat r5-r1 output supplementary.pdf

\end{document}